\documentclass[runningheads]{llncs}
\usepackage{xcolor}

\usepackage[T1]{fontenc}
\usepackage{graphicx}
\usepackage{twemojis}

\usepackage{amsmath} 
\usepackage{todonotes}

\usepackage{tabularx}
\usepackage{booktabs}
\usepackage[colorlinks=true, linkcolor=blue, citecolor=blue, urlcolor=blue]{hyperref}

\usepackage{color}

\usepackage{multirow}
\usepackage{cite}
\usepackage{amssymb}

\begin{document}

\title{OSCAR: Optical-aware Semantic Control for Aleatoric Refinement in Sar-to-Optical Translation}
%
%

\author{
Hyunseo Lee\inst{1} \and
Sang Min Kim\inst{1} \and
Ho Kyung Shin\inst{1} \and
Taeheon Kim\inst{2} \and 
Woo-Jeoung Nam\inst{1}\thanks{Corresponding author.}
}

\authorrunning{H. Lee et al.}

\institute{
\textsuperscript{1}Kyungpook National University \\
\textsuperscript{2}Korea Aerospace Research Institute \\
\email{
\{heart2002101, klasskoch, tlsghrud\}@knu.ac.kr, honey25@kari.re.kr nwj0612@knu.ac.kr \\
}
}
\maketitle              

\begin{figure*}[!h]
    \centering
    \includegraphics[width=1.0\textwidth]{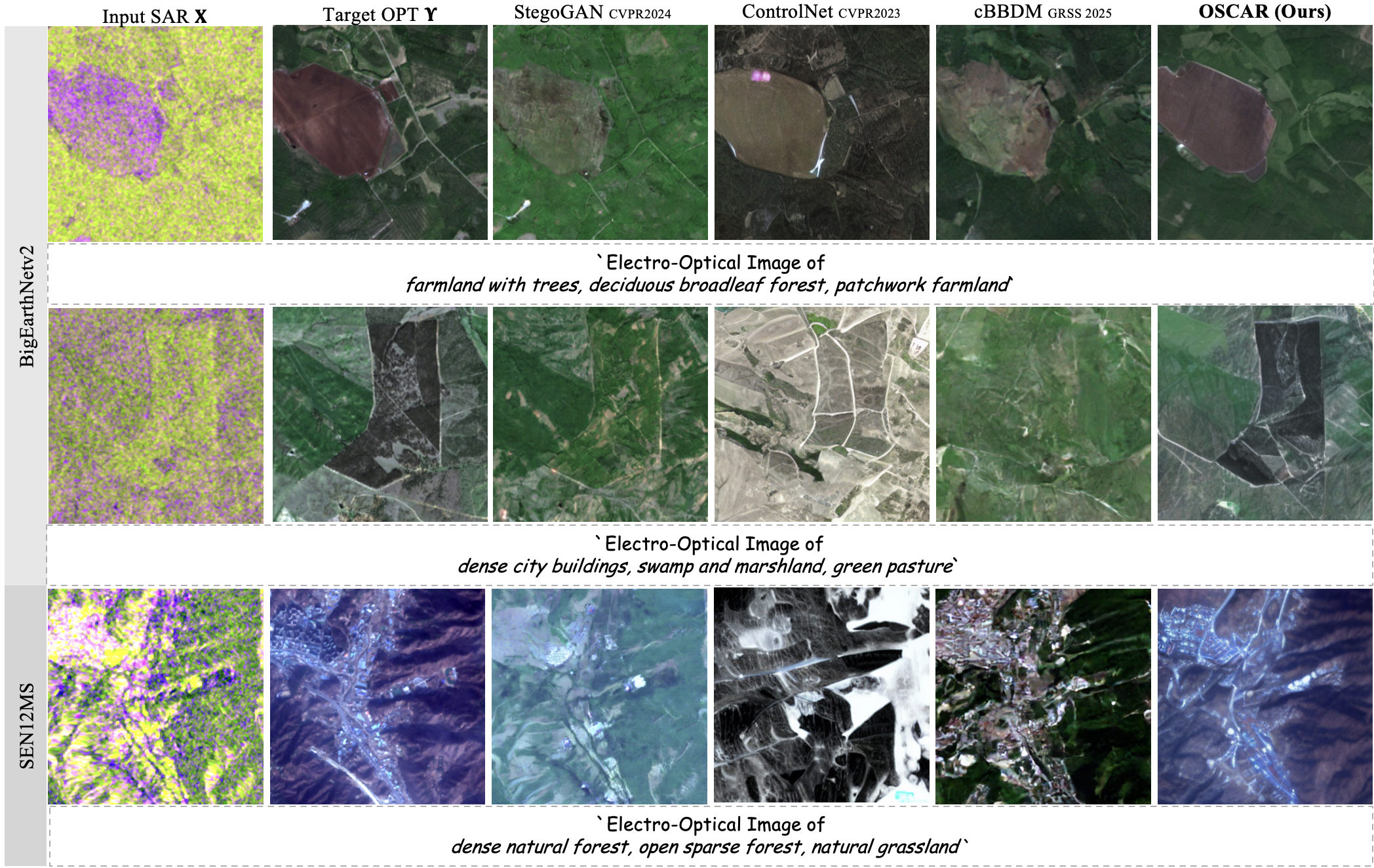}
    \caption{Qualitative comparison of S2O translation results on three datasets. We compare our proposed method against state-of-the-art baselines (StegoGAN, ControlNet, cBBDM). Below each example, the corresponding Class-aware Prompt generated by our Optical-aware SAR Encoder is displayed. As shown, our method demonstrates superior structural fidelity and perceptual quality compared to other state-of-the-art models, effectively preserving semantic details while suppressing artifacts.} 
    \label{fig:qualitative_results}
\end{figure*}

\begin{abstract}

Synthetic Aperture Radar (SAR) provides robust all-weather imaging capabilities; however, translating SAR observations into photo-realistic optical images remains a fundamentally ill-posed problem. Current approaches are often hindered by the inherent speckle noise and geometric distortions of SAR data, which frequently result in semantic misinterpretation, ambiguous texture synthesis, and structural hallucinations. To address these limitations, a novel SAR-to-Optical (S2O) translation framework is proposed, integrating three core technical contributions: (i) Cross-Modal Semantic Alignment, which establishes an Optical-Aware SAR Encoder by distilling robust semantic priors from an Optical Teacher into a SAR Student (ii) Semantically-Grounded Generative Guidance, realized by a Semantically-Grounded ControlNet that integrates class-aware text prompts for global context with hierarchical visual prompts for local spatial guidance; and (iii) an Uncertainty-Aware Objective, which explicitly models aleatoric uncertainty to dynamically modulate the reconstruction focus, effectively mitigating artifacts caused by speckle-induced ambiguity. Extensive experiments demonstrate that the proposed method achieves superior perceptual quality and semantic consistency compared to state-of-the-art approaches. Our project page is available at \url{https://eunoiahyunseo.github.io/OSCAR/}

\keywords{SAR-to-Optical Translation \and Sematic-Guidance} 
\end{abstract}

\section{Introduction}

Synthetic Aperture Radar (SAR) is indispensable for all-weather, day-and-night remote sensing, overcoming the limitations of Electro-Optical (EO) sensors hindered by cloud cover or illumination~\cite{yamaguchiDisasterMonitoringFully2012, gelautzSARImageSimulation1998}. However, SAR's unique scattering properties create a significant domain gap with optical data, hindering human interpretation and the direct application of standard vision algorithms. To bridge this, SAR-to-Optical (S2O) translation has emerged as a key research direction~\cite{heDOGANDINOBasedOpticalPriorDriven2025, liBBDMImagetoimageTranslation2023, zhangAddingConditionalControl2023}. Recent advancements in generative models, particularly GANs and DDPMs, show promise in synthesizing realistic optical textures from radar signals, thereby extending SAR's utility for downstream tasks~\cite{wuStegoGANLeveragingSteganography2024, zhuUnpairedImagetoImageTranslation2020}.

Despite these advancements, S2O translation remains a fundamentally ill-posed problem. The inherent speckle noise and geometric distortions in SAR images create severe domain discrepancies, frequently resulting in spatial misalignment, high-frequency hallucinations, and over-smoothed textures. As illustrated in Fig.~\ref{fig:qualitative_results}, conventional ControlNet-based approaches \cite{zhangAddingConditionalControl2023} often suffer from structural hallucinations and artifacts due to speckle noise. To overcome this, we propose an architecture that enhances structural control by injecting pixel-level semantic representations, inspired by PASD \cite{yangPixelAwareStableDiffusion2024}. This allows our model to utilize semantic priors as a buffer against radar-specific noise, ensuring semantically accurate and texturally realistic synthesis.

To realize this vision, a novel framework is introduced, centered on an Optical-Aware SAR Encoder designed to extract the 'richer, optical-aligned semantic information'. Since SAR data inherently lacks color and spectral information, training a semantic extractor solely on radar signals is insufficient for capturing optical-like semantics. To address this, a Cross-Modal Semantic Alignment framework is employed. DINOv3-SAT \cite{simeoniDINOv32025}, a powerful large-scale vision foundation model pre-trained on satellite imagery, is utilized as the teacher network. The SAR encoder is trained to mimic the teacher's feature representation, thereby aligning the SAR feature space with the rich semantic manifold of the optical domain \cite{bardesVICRegVarianceInvarianceCovarianceRegularization2022, kangAdaptiveClassToken2024, hintonDistillingKnowledgeNeural2015}. This alignment ensures that the encoder extracts robust, modality-agnostic representations that encapsulate the visual logic of the scene—effectively implementing the optical-aware guiding strategy.

Building upon these learned representations, a Semantically-Grounded ControlNet is proposed to perform the actual guidance within the compressed latent space. Instead of using the encoder merely for classification, its hierarchical intermediate features are extracted and injected into the ControlNet as visual prompts. Echoing the pixel-aware cross-attention mechanism of PASD \cite{yangPixelAwareStableDiffusion2024}, these features provide the diffusion model with dense, spatially-grounded semantic guidance. Crucially, unlike raw structural cues, these optical-aligned semantic features act as spatial anchors within the latent manifold, stabilizing the generation process against geometric shifts. Simultaneously, Class-aware Text Prompts are generated based on the encoder's high-confidence predictions. These prompts serve to establish the global semantic tone (e.g., domain style).  This synergy between class-aware text prompts and optical-aligned semantic features enables the synthesis of images that are both stylistically coherent and topologically accurate, effectively overcoming the limitations of previous unguided methods.

Finally, even with semantic guidance, the intrinsic ambiguity \cite{kendallWhatUncertaintiesWe2017} caused by speckle noise remains a challenge. Forcing the model to reconstruct every pixel with equal importance can lead to overfitting to noise, resulting in the artificial high-frequency artifacts mentioned earlier. To address this, an Uncertainty-Aware Objective is incorporated that explicitly models aleatoric uncertainty. By estimating a pixel-wise confidence map, the reconstruction loss is dynamically modulated, down-weighting the penalty in regions with high uncertainty. This mechanism prevents the model from hallucinating details in ambiguous areas while encouraging sharp and accurate reconstruction in high-confidence regions, leading to more perceptually realistic outcomes.

In summary, the main contributions are as follows: \begin{itemize} \item An Optical-Aware SAR Encoder trained via DINOv3-based knowledge distillation is proposed, which aligns SAR representations with optical semantics to provide robust, modality-agnostic features. \item A Semantically-Grounded ControlNet is introduced, injecting hierarchical visual prompts into the latent diffusion process, effectively injecting pixel-level semantic priors into the structural guidance process. \item An Uncertainty-based Objective is implemented to explicitly handle aleatoric uncertainty, significantly reducing hallucinations and improving the perceptual quality of the generated images. \end{itemize}

\section{Related Work}

\subsection{S2O Image Translation}

\subsubsection{GAN-based Methods} 
Early S2O translation primarily utilized GANs, following either the Pix2Pix framework for paired translation with pixel-wise supervision~\cite{isolaImagetoImageTranslationConditional2018, zuoSARtoOpticalImageTranslation2021a} or the CycleGAN architecture for unpaired translation via cycle-consistency constraints~\cite{wangSARtoOpticalImageTranslation2019, yangFGGANFineGrainedGenerative2022}. To further mitigate modality-induced information loss, StegoGAN~\cite{wuStegoGANLeveragingSteganography2024} introduced steganographic constraints to embed and preserve semantic features during the translation process.

\subsubsection{Diffusion-based Methods}
Diffusion models have recently outperformed GANs in S2O tasks by offering superior stability and textural realism through progressive denoising~\cite{hoDenoisingDiffusionProbabilistic2020, sahariaPaletteImagetoImageDiffusion2022, baiConditionalDiffusionSAR2024, qinConditionalDiffusionModel2024}. Notably, bridge-based frameworks such as BBDM~\cite{liBBDMImagetoimageTranslation2023} and its conditional variant cBBDM~\cite{kimConditionalBrownianBridge2025} model the direct trajectory between SAR and optical distributions. By formulating the translation as a bidirectional bridge process, these methods achieve more precise modality conversion than standard pixel-space diffusion models.

\subsection{Controllable and Semantic-Guided Generation}

\subsubsection{Controllable Diffusion Models} ControlNet enables spatial controllability by adding a trainable conditioning branch to a frozen diffusion backbone \cite{zhangAddingConditionalControl2023}. To handle degraded inputs, PASD \cite{yangPixelAwareStableDiffusion2024} employs a degradation-removal branch to extract multi-scale structural features and injects them into the U-Net via a pixel-aware cross-attention (PACA) module. This architecture allows for the enforcement of structural fidelity even when input observations are contaminated by noise or blur.

\subsubsection{Foundation Models as a Semantic Extractors}
Vision foundation models like DINOv3 provide robust semantic representations through large-scale self-supervised pretraining \cite{simeoniDINOv32025}. A significant advancement in this field is DINOv3-SAT, which is specifically pretrained on the SAT-493M dataset—a massive-scale collection of 493 million satellite images. This extensive pretraining on overhead imagery allows the model to capture domain-specific geometric patterns and scene statistics more effectively than general web-scale encoders.

In the context of S2O translation, DOGAN \cite{heDOGANDINOBasedOpticalPriorDriven2025} recently demonstrated the utility of integrating DINO-based priors to provide semantic guidance during the generative process. Our framework extends this approach by employing DINOv3-SAT to leverage its native alignment with satellite-specific geometry, ensuring the extracted semantics are optimized for remote sensing scenes.

\section{Proposed Method}
\subsection{Overall Framework}
To address the inherent ill-posedness of S2O translation—characterized by speckle noise and geometric ambiguity—OSCAR (Optical-aware Semantic Control for Aleatoric Refinement) is proposed. As illustrated in Fig.~\ref{fig:distill}, Fig.~\ref{fig:framework}, the framework consists of two sequential stages. First, in the Cross-Modal Semantic Alignment stage (Sec.~\ref{sec:rep_learning}), an Optical-Aware SAR Encoder is constructed via Cross-Modal Distillation. This encoder learns to extract robust, optical-aligned semantic representations from noisy SAR inputs by mimicking a pre-trained Optical Teacher. Second, in the Semantically-Grounded Generative Guidance stage (Sec.~\ref{sec:translation}), a Semantically-Grounded ControlNet is employed. This generative model utilizes the learned representations as Hierarchical Visual Prompts for spatially-grounded semantic guidance and Class-aware Text Prompts for global context. Finally, to mitigate artifacts in ambiguous regions, an Uncertainty-Aware Objective is introduced to dynamically modulate the reconstruction focus.

\subsection{Cross-Modal Semantic Alignment}
\label{sec:rep_learning}
Since SAR data lacks color and spectral information, training a semantic extractor solely on radar signals is insufficient for capturing optical-like semantics. To bridge this gap, a Cross-Modal Distillation framework is adopted to transfer the rich semantic priors of a vision foundation model to the SAR domain.

\subsubsection{Optical-Aware SAR Encoder}

To extract robust semantic representations, DINOv3-SAT \cite{simeoniDINOv32025}, pre-trained on satellite imagery, serves as the foundational backbone for both the Optical Teacher (T) and the SAR Student (S). To adapt this large-scale model to the distinct modalities while ensuring parameter efficiency, Low-Rank Adaptation (LoRA)~\cite{huLoRALowRankAdaptation2021} is integrated into both networks. LoRA is applied to the attention matrices ($W_{q,k,v,o}$), and a linear classification head is attached to the final [CLS] token to predict multi-label land-cover classes.

\begin{figure}[t]
    \centering
    \includegraphics[width=1.0\textwidth]{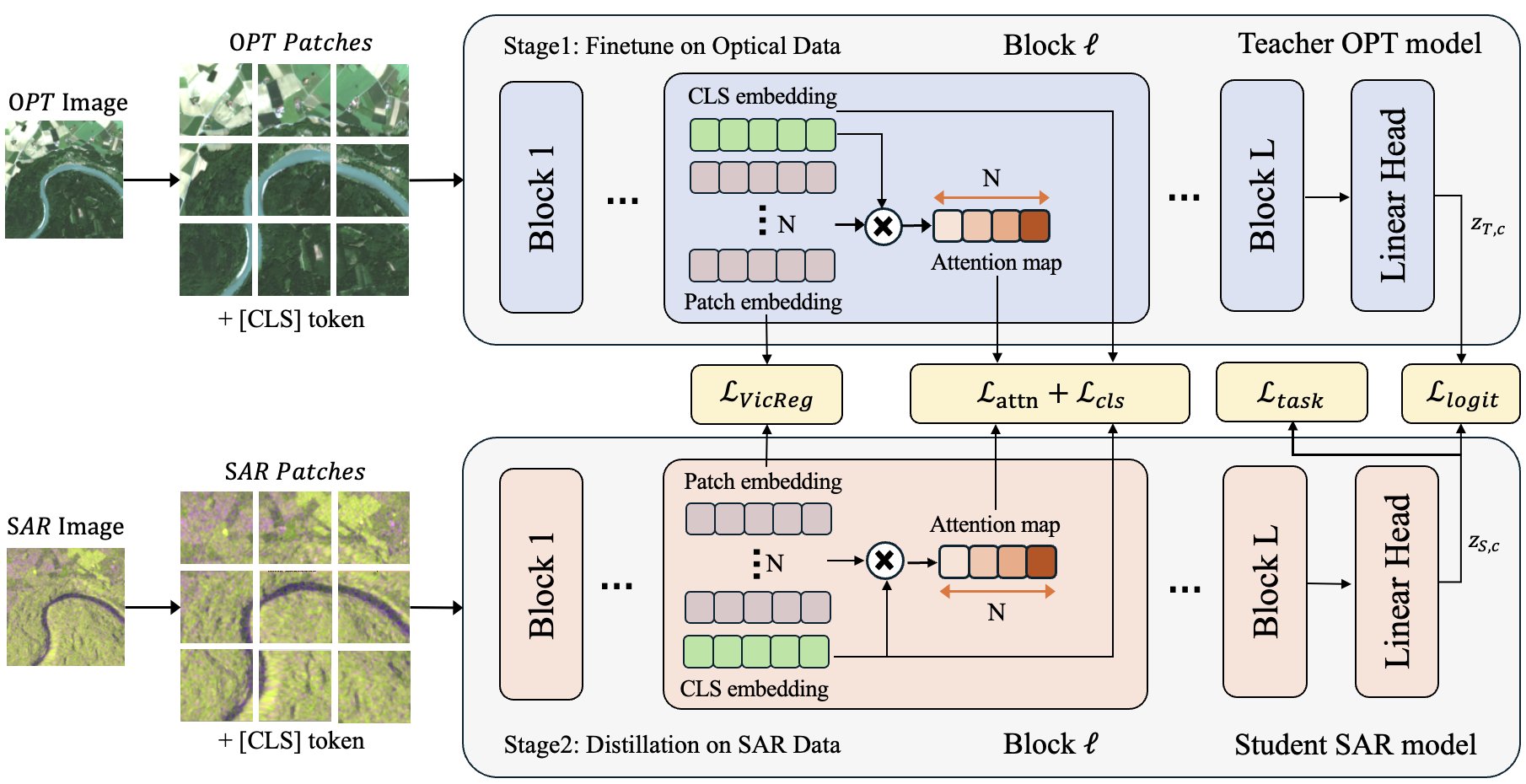}
    \caption{Schematic of the proposed Cross-Modal Semantic Alignment. A Cross-Modal distillation strategy is employed where the trainable SAR Student mimics the frozen Optical Teacher.}
    \label{fig:distill}
\end{figure}

Building upon this shared architecture, the semantic alignment proceeds in two stages as illustrated in Fig.~\ref{fig:distill}. First, the Optical Teacher is supervisedly fine-tuned on the optical imagery to establish a reliable semantic oracle. Subsequently, while freezing the Teacher, the SAR Student is trained to mimic the Teacher's feature representations through a multi-level distillation strategy comprising logit-level~\cite{hintonDistillingKnowledgeNeural2015}, feature-level~\cite{bardesVICRegVarianceInvarianceCovarianceRegularization2022}, and structural alignment~\cite{kangAdaptiveClassToken2024}.

To achieve this comprehensive alignment, the total objective function for the Student network, $\mathcal{L}_{total}$, is defined as:
\begin{equation}
\mathcal{L}_{total} = \mathcal{L}_{task} + \lambda_{kd}\mathcal{L}_{logit} + \lambda_{attn}(\mathcal{L}_{attn} + \mathcal{L}_{cls}) + \mathcal{L}_{VicReg}
\end{equation}
The optimization is driven by four specific components, each addressing a distinct aspect of the alignment strategy:

(i) \textbf{Class-Balanced Supervised Task Loss ($\mathcal{L}_{task}$):} To address the long-tail distribution inherent in land-cover datasets \cite{clasenReBENRefinedBigEarthNet2025, schmittSEN12MSCuratedDataset2019}, we employ a weighted Binary Cross-Entropy loss \cite{cuiClassBalancedLossBased2019}. The task loss is formulated as:
\begin{equation}
\mathcal{L}_{task} = - \frac{1}{C} \sum_{c=1}^{C} \left[ w_c \cdot y_c \log(\sigma(z_{S,c})) + (1-y_c) \log(1-\sigma(z_{S,c})) \right]
\end{equation}
where $C$ denotes the total number of land-cover classes, $\sigma$ is sigmoid function and $w_c$ is the positive class weight inversely proportional to class frequency:
\begin{equation}
    w_c = \text{clip}\left(\frac{N - N_c}{N_c}, \alpha_{min}, \alpha_{max}\right)
\end{equation}
Here, $N$ is the total number of samples and $N_c$ is the count for class $c$, with clipping thresholds $\alpha_{min}$ and $\alpha_{max}$ utilized to stabilize training.

(ii) \textbf{Logit-level Distillation ($\mathcal{L}_{logit}$):} To transfer soft semantic probability distributions from the teacher, we apply soft-target distillation with a temperature parameter $\mathcal{T}$:
\begin{equation}
\mathcal{L}_{logit} = - \frac{1}{C} \sum_{c=1}^{C} \left[ p_{T,c} \log(p_{S,c}) + (1-p_{T,c}) \log(1-p_{S,c}) \right]
\end{equation}
where $p_{S,c} = \sigma(z_{S,c} / \mathcal{T})$ and $p_{T,c} = \sigma(z_{T,c} / \mathcal{T})$ denote the soft probabilities of the student and teacher, respectively.

(iii) \textbf{Intermediate Feature Alignment ($\mathcal{L}_{attn}, \mathcal{L}_{cls}$):} 
Moving beyond class probabilities, we enforce the synchronization of both global semantic tokens and local spatial attention maps across a specific subset of deeper layers $\mathcal{L}_{d} = \{11, 14, 17, 20, 23\}$. Since the DINOv3-SAT (ViT-L/16) architecture consists of 24 transformer blocks, we intentionally align these layers to focus on distilling high-level semantic representations and global context. Let $A^l \in \mathbb{R}^{1 \times N}$ be the attention map and $cls^l \in \mathbb{R}^{d}$ be the feature vector of the [CLS] token at layer $l$. The structural losses are defined as:
\begin{equation}
\mathcal{L}_{attn} = \frac{1}{|\mathcal{L}_d|} \sum_{l \in \mathcal{L}_d} \sum_{n=1}^{N} A_{T,n}^l \log \left( \frac{A_{T,n}^l}{A_{S,n}^l} \right), \quad \mathcal{L}_{cls} = \frac{1}{|\mathcal{L}_d|} \sum_{l \in \mathcal{L}_d} \| cls_S^l - cls_T^l \|_2^2
\end{equation}
This ensures the SAR encoder focuses on the same spatial regions and extracts similar semantic concepts as the optical teacher. This ensures the SAR encoder focuses on the same spatial regions as the optical teacher. While we utilize identical architectures, additional adapter layers could be employed for alignment if the student and teacher architectures differed.

(iv) \textbf{VicReg Regularization ($\mathcal{L}_{VicReg}$):} Finally, to prevent feature collapse and enhance representation diversity, we adapt the VicReg objective \cite{bardesVICRegVarianceInvarianceCovarianceRegularization2022}, incorporating invariance, variance, and covariance terms:
\begin{equation}
\mathcal{L}_{VicReg} = \lambda_{inv}\mathcal{L}_{inv} + \mu_{var}\mathcal{L}_{var} + \nu_{cov}\mathcal{L}_{cov}
\end{equation}

\begin{figure}[t]
    \centering
    \includegraphics[width=1.0\textwidth]{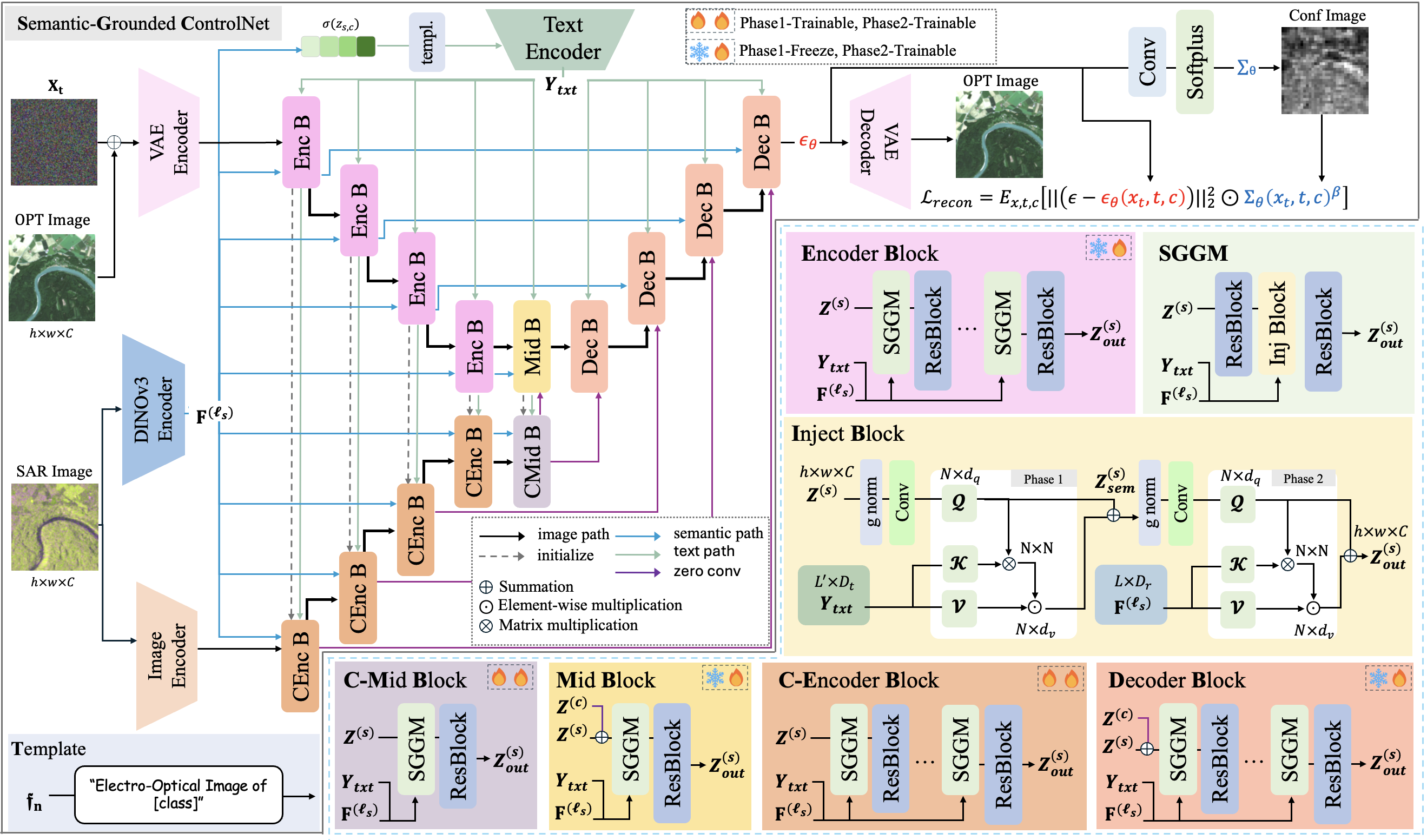}
    \caption{Overview of the proposed OSCAR framework. The architecture utilizes the Student SAR model (pre-trained in Fig.~\ref{fig:distill}) as the Optical-aware SAR Encoder. Key components include the injection of hierarchical semantic information extracted by the Student SAR model, class-aware text prompts for global semantic context, and the estimation of a pixel-wise confidence map to optimize the uncertainty-aware objective.}
    \label{fig:framework}
\end{figure}

\subsection{Semantically-Grounded Generative Guidance}
\label{sec:translation}
Building upon the optical-aligned representations, a Semantically-Grounded ControlNet is proposed, as illustrated in Fig.~\ref{fig:framework}. The introduction highlighted critical challenges, including speckle noise, geometric distortions, and the ill-posed nature of S2O mapping. To address these limitations, explicit semantic guidance is injected through two complementary pathways. Class-aware Text Prompts are employed to establish the global semantic context, while Hierarchical Visual Prompts are utilized for local spatial grounding.

\subsubsection{Class-aware Text Prompts}
To provide the generation process with global semantic context, dynamic text prompts are synthesized using the pre-trained Optical-aware SAR Encoder. Sigmoid probabilities $p = \sigma(z_{S,c})$ are computed, and the top-$k$ classes exceeding a high confidence threshold $\tau$ are selected:
\begin{equation}
    \mathcal{C}_{active} = \{ c_i \mid p_i > \tau, \ i \in \text{Top-}k(p) \}
\end{equation}
These classes are formatted into a prompt template:
\begin{equation}
    \text{Prompt} = \text{``Electro-Optical Image of } [\mathcal{C}_{active}]\text{''}
\end{equation}
This explicitly sets the global semantic tone via Class-aware Text Prompts, reducing domain ambiguity.

\subsubsection{Semantically-Grounded Guidance Module}
While text prompts provide global context, they lack spatial precision. To ensure both semantic coherence and pixel-level structural fidelity, we introduce a Semantically-Grounded Guidance Module (SGGM). This module integrates conditions in two progressive phases: injecting global context first, followed by local spatial grounding.

Hierarchical Visual Prompts are extracted from the previously aligned layer set $\mathcal{L}_d = \{11, 14, 17, 20, 23\}$. Specifically, we define a subset $\mathcal{L}_h = \{14, 17, 20, 23\}$ by selecting the four deepest layers from $\mathcal{L}_d$ to provide robust, high-level semantic guidance. These features, forming the set $\mathcal{F}_{visual} = \{ F^{(l)} \mid l \in \mathcal{L}_h \}$, are mapped to U-Net stages $s$ via the following strategy:
\begin{equation}
    l_s = 
    \begin{cases} 
    14, 17, 20, 23 & \text{for Encoder stages } s \in \{1, 2, 3, 4\} \\
    23 & \text{for Middle block } s_{mid} \\
    23, 20, 17, 14 & \text{for Decoder stages } s \in \{1, 2, 3, 4\}
    \end{cases}
\end{equation}
By utilizing these deeper semantic layers from $\mathcal{L}_h$, the model stabilizes the generation process against the inherent noise in SAR observations, ensuring that each stage of the diffusion process is grounded in a modality-aligned semantic manifold.
At each stage, the guidance is applied sequentially:

\textbf{Phase 1} First, the intermediate feature $Z^{(s)}$ acts as the query, while the Class-aware Text Prompts $Y_{txt}$ serve as (Keys/Values). This step injects the global domain context (e.g., land-cover type) into the feature space:
\begin{equation}
    Z_{sem}^{(s)} = Z^{(s)} + \text{Softmax}\left( \frac{(Z^{(s)}W_Q)(Y_{txt}W_K)^\top}{\sqrt{d_k}} \right) (Y_{txt}W_V)
\end{equation}

\textbf{Phase 2} Subsequently, the semantically enriched feature $Z_{sem}^{(s)}$ serves as the query for the Hierarchical Visual Prompts $F^{(l_s)}$ (Keys/Values). This enforces pixel-level semantic alignment with the optical manifold:
\begin{equation}
    Z_{out}^{(s)} = Z_{sem}^{(s)} + \text{Softmax}\left( \frac{(Z_{sem}^{(s)}W'_Q)(F^{(l_s)}W'_K)^\top}{\sqrt{d_k}} \right) (F^{(l_s)}W'_V)
\end{equation}
This sequential design ensures that the generated textures are first semantically conditioned by the text and then spatially instantiated with dense semantic details by the visual prompts.

\subsection{Uncertainty-Aware Objective}
\label{sec:uncertainty}
Standard MSE loss treats all pixels equally, often forcing the model to hallucinate high-frequency artifacts in regions dominated by SAR speckle noise. This issue is addressed by explicitly modeling aleatoric uncertainty \cite{kendallWhatUncertaintiesWe2017}.

The model predicts a pixel-wise confidence map $\Sigma_\theta(x_t, t, c)$ alongside the noise $\epsilon_\theta$. $\Sigma_\theta$ represents the estimated inverse variance of the prediction. The objective is formulated as the negative log-likelihood of a Gaussian distribution:
\begin{equation}
    \mathcal{L}_{total} = \mathcal{L}_{recon} + \lambda \mathcal{L}_{reg}
\end{equation}

The Uncertainty-weighted Reconstruction Loss is defined as:
\begin{equation}
    \mathcal{L}_{recon} = \mathbb{E}_{x, t, c} \left[ \frac{1}{2} \| (\epsilon - \epsilon_\theta(x_t, t, c)) \|_2^2 \odot \Sigma_\theta(x_t, t, c)^\beta \right]
\end{equation}
Here, $\Sigma_\theta$ acts as a dynamic attenuation mechanism. In regions with high aleatoric uncertainty (e.g., severe speckle or ambiguous shadows), the model learns to predict a lower $\Sigma_\theta$, thereby down-weighting the reconstruction penalty. This prevents the model from overfitting to noise.

To prevent the trivial solution where $\Sigma_\theta \to 0$, the Regularization Loss serves as the log-determinant constraint:
\begin{equation}
    \mathcal{L}_{reg} = \mathbb{E}_{x, t, c} \left[ - \log(\Sigma_\theta(x_t, t, c)^\beta + \delta) \right]
\end{equation}
This objective encourages high-fidelity generation in confident regions while suppressing artifacts in ambiguous areas.

\begin{figure*}[t]
    \centering
    \includegraphics[width=1.0\textwidth]{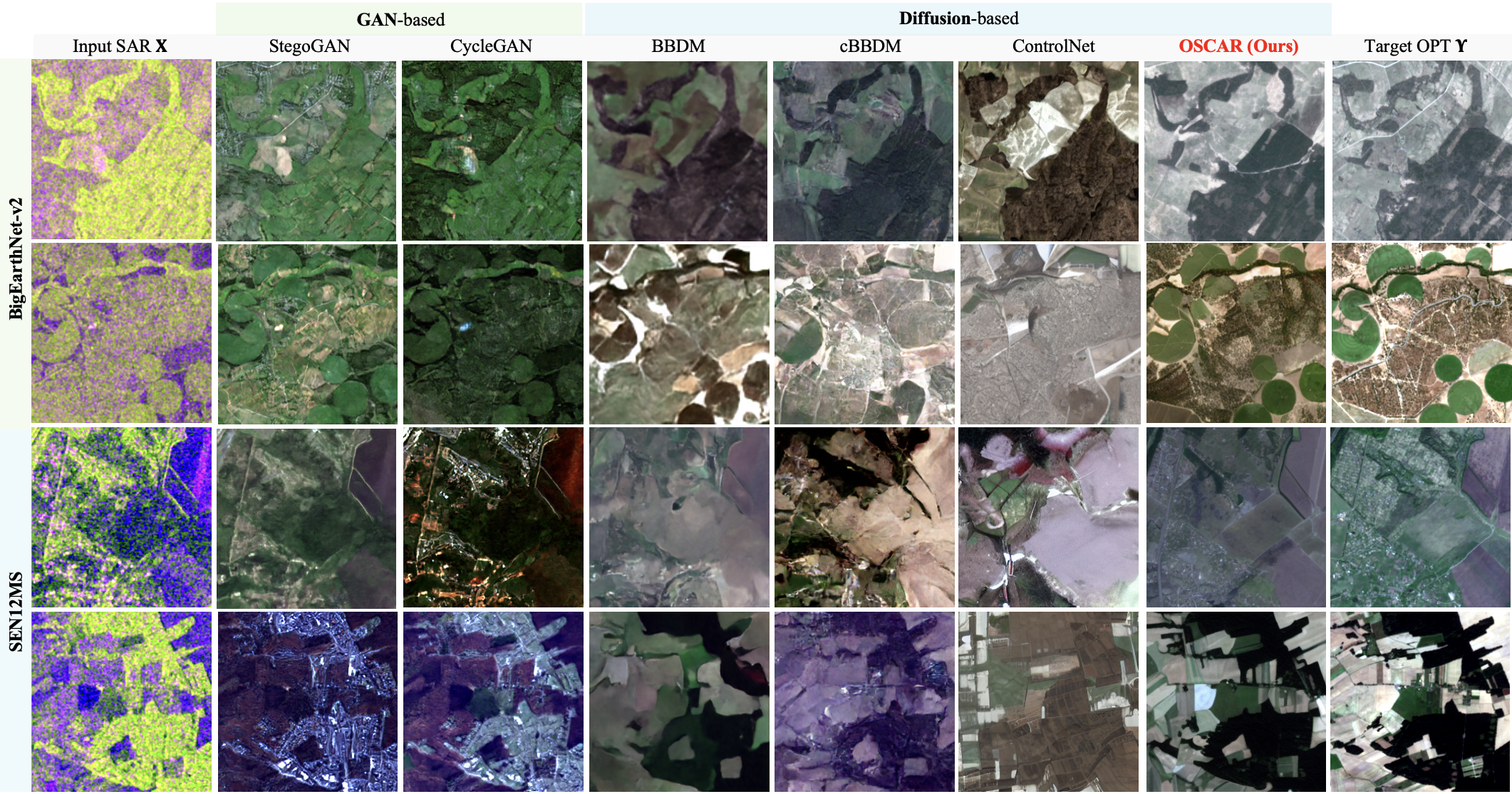}
    \caption{Qualitative comparison on BigEarthNet-v2 and SEN12MS. While baseline methods often suffer from geometric distortion or texture blurring, OSCAR generates structurally coherent and photorealistic results.}
    \label{fig:prompt_vis}
\end{figure*}

\section{Experiments}

\subsection{Datasets}

\subsubsection{BigEarthNet-v2}
BigEarthNet-v2 \cite{clasenReBENRefinedBigEarthNet2025} serves as our primary regional benchmark, comprising approximately 590,410 pairs of SAR and optical patches across diverse European landscapes. We utilize Sentinel-1 VV and VH polarization data as the structural input, while Sentinel-2 RGB bands serve as the optical ground truth. Additionally, the multi-label annotations from the CORINE Land Cover database \cite{europeanenvironmentagencyCORINELandCover2019} are leveraged to train the auxiliary classifier for class-aware prompting.

\subsubsection{SEN12MS}
To validate global generalization, we employ the SEN12MS dataset \cite{schmittSEN12MSCuratedDataset2019}. This benchmark extends the geographic scope worldwide, providing pairs of Sentinel-1 and Sentinel-2 imagery alongside MODIS Land Cover maps \cite{friedlGlobalLandCover2002}. We constructed a balanced subset of 40,000 pairs by randomly sampling 10,000 triplets from each of the four seasons. This configuration facilitates robust training and evaluation under distinct seasonal variations and global landscape distributions, while demonstrating the model's robustness even in data-constrained environments.

\subsection{Experimental Results}

\begin{table*}[t]
\caption{Quantitative comparison of S2O translation results. Methods are categorized by generative architecture (GAN vs. Diffusion).}
\label{tab:quantitative_results_grouped}
\centering
\resizebox{\textwidth}{!}{
\setlength{\tabcolsep}{3pt}
\begin{tabular}{l l | cccccc ccc}
\toprule
\textbf{Type} & \textbf{Method} 
& \textbf{DISTS}~$\downarrow$ & \textbf{LPIPS}~$\downarrow$ & \textbf{SAM}~$\downarrow$ & \textbf{FID}~$\downarrow$ & \textbf{KID}~$\downarrow$ & \textbf{$D_{\lambda}$}~$\downarrow$ 
& \textbf{SSIM}~$\uparrow$ & \textbf{SCC}~$\uparrow$ & \textbf{IS}~$\uparrow$ \\
\midrule

\multicolumn{11}{c}{Dataset: BigEarthNet-v2~\cite{clasenReBENRefinedBigEarthNet2025}} \\
\midrule
\multirow{2}{*}{\rotatebox{90}{GAN}}
& CycleGAN \cite{zhuUnpairedImagetoImageTranslation2020}
& 0.3559 & 0.4944 & 0.2570 & 146.81 & 0.0974 & 0.1200 
& 0.1269 & 0.0132 & 2.9406 \\
& StegoGAN \cite{wuStegoGANLeveragingSteganography2024}
& 0.2897 & 0.4127 & 0.1939 & 120.01 & 0.0714 & 0.0852 
& 0.3208 & 0.0203 & 2.7211 \\
\cmidrule{1-11}
\multirow{4}{*}{\rotatebox{90}{Diffusion}}
& ControlNet \cite{zhangAddingConditionalControl2023}
& 0.3545 & 0.5643 & 0.2687 & 103.16 & 0.0362 & 0.1393 
& 0.1609 & 0.0002 & 3.5776 \\
& BBDM \cite{liBBDMImagetoimageTranslation2023}
& 0.2818 & 0.4020 & 0.2085 & 83.50  & 0.0412 & 0.1094 
& 0.3542 & 0.0186 & 2.7358   \\
& cBBDM \cite{kimConditionalBrownianBridge2025}
& 0.2687 & 0.4021 & 0.2003 & 95.66  & 0.0509 & 0.0860
& 0.3326 & 0.0105 & 3.0439 \\
& \textbf{OSCAR (Ours)}
& \textbf{0.2404} & \textbf{0.3831} & \textbf{0.1638} & \textbf{56.32} & \textbf{0.0104} & \textbf{0.0808}
& \textbf{0.3564} & \textbf{0.0276} & \textbf{4.4775} \\

\midrule
\multicolumn{11}{c}{Dataset: SEN12MS~\cite{schmittSEN12MSCuratedDataset2019}} \\
\midrule
\multirow{2}{*}{\rotatebox{90}{GAN}}
& CycleGAN~\cite{zhuUnpairedImagetoImageTranslation2020}
& 0.4436 & 0.7140 & 0.5158 & 75.77 & 0.0364 & 0.2179 
&    0.0827  & 0.0004 & 3.7160 \\
& StegoGAN~\cite{wuStegoGANLeveragingSteganography2024}
& 0.4303 & 0.7146 & 0.4672 & 69.69 & 0.0350 & 0.2133 
& 0.1143      & 0.0008 & 3.4675 \\
\cmidrule{1-11}
\multirow{4}{*}{\rotatebox{90}{Diffusion}}
& ControlNet~\cite{zhangAddingConditionalControl2023}
& 0.4546 & 0.7360 & 0.3998 & 71.08 & 0.0344 & 0.2023 
& 0.1162     & 0.0010 & 3.2797 \\
& BBDM~\cite{liBBDMImagetoimageTranslation2023}
& 0.3931 & 0.6181 & 0.4117 & 120.83 & 0.0644 & 0.1753 
& 0.2445  & 0.0084 & 3.1529 \\
& cBBDM~\cite{kimConditionalBrownianBridge2025}
& 0.3311 & 0.5191 & 0.3513 & 71.33 & 0.0382 & 0.1616 
& 0.2553 & 0.0047 & 3.6027 \\
& \textbf{OSCAR (Ours)}
& \textbf{0.2710} & \textbf{0.4362} & \textbf{0.1923} & \textbf{34.73} & \textbf{0.0108} & \textbf{0.1079} 
& \textbf{0.3625} & \textbf{0.0099} & \textbf{5.1869} \\
\bottomrule
\end{tabular}
}
\end{table*}

\subsubsection{Quantitative Evaluation.}
We compare the proposed framework against state-of-the-art GAN-based \cite{zhuUnpairedImagetoImageTranslation2020, wuStegoGANLeveragingSteganography2024} and Diffusion-based approaches \cite{zhangAddingConditionalControl2023, liBBDMImagetoimageTranslation2023, kimConditionalBrownianBridge2025}, with quantitative results summarized in Table~\ref{tab:quantitative_results_grouped}. As evidenced, OSCAR establishes a new state-of-the-art across all metric categories on both regional (BigEarthNet-v2) and global (SEN12MS) benchmarks, significantly outperforming existing baselines. In terms of distribution matching and perceptual quality, our method demonstrates drastic improvements. Specifically, compared to the best performing baselines, OSCAR reduced the FID score by 32.5\% on BigEarthNet-v2 (vs. BBDM) and by over 50.2\% on SEN12MS (vs. StegoGAN). Similarly, the Inception Score (IS) increased by 25.1\% and 39.6\% on each dataset, respectively. These margins confirm that our model synthesizes significantly more realistic and diverse optical textures, effectively bridging the domain gap where previous methods struggled. Regarding structural and spectral fidelity, our framework excels in preserving the underlying terrain layout while minimizing color distortion. Unlike conventional generative models that often trade structural accuracy for textural richness, OSCAR achieves the best performance in SSIM and SCC. Notably, on the challenging SEN12MS dataset, our method improved SSIM by 42.0\% compared to cBBDM. Furthermore, the lowest spectral distortion ($D_{\lambda}$) and SAM scores across both datasets validate that the synergy of class-aware text prompts and hierarchical visual prompts effectively guides the generation, ensuring both topological accuracy and spectral consistency.

\begin{table*}[t]
\caption{Ablation study on BigEarthNet-v2. We analyze the impact of cross-modal alignment (Aln.), hierarchical visual prompts (Hier.), and class-aware text prompts (Text).}
\label{tab:ablation_total}
\centering
\resizebox{1.0\textwidth}{!}{
\begin{tabular}{c ccc cccccc ccc}
\toprule
\textbf{ID} & \textbf{Aln.} & \textbf{Hier.} & \textbf{Text} & \textbf{DISTS}~$\downarrow$ & \textbf{LPIPS}~$\downarrow$ & \textbf{SAM}~$\downarrow$ & \textbf{FID}~$\downarrow$ & \textbf{KID}~$\downarrow$ & \textbf{$D_{\lambda}$}~$\downarrow$ & \textbf{SSIM}~$\uparrow$ & \textbf{SCC}~$\uparrow$ & \textbf{IS}~$\uparrow$ \\
\midrule

(i)  & -- & \checkmark & \checkmark & 0.2413 & 0.3839 & 0.1649 & \textbf{55.78} & \textbf{0.0097} & 0.0785 & 0.3553 & 0.0256 & 4.4775 \\
\midrule
(ii) & \checkmark & --         & --         & 0.2438 & 0.3883 & 0.1682 & 57.76 & 0.0110 & 0.0810 & 0.3493 & 0.0252 & 4.4773 \\
(iii) & \checkmark & --         & \checkmark & 0.2433 & 0.3933 & 0.1736 & 58.60 & 0.0113 & 0.0831 & 0.3330 & 0.0241 & 4.4775 \\
(iv) & \checkmark & \checkmark & --         & 0.2425 & 0.3862 & 0.1646 & 57.34 & 0.0105 & 0.0809 & \textbf{0.3572} & 0.0275 & 4.4775 \\
\midrule
(v)  & \checkmark & \checkmark & \checkmark & \textbf{0.2404} & \textbf{0.3831} & \textbf{0.1638} & 56.32 & 0.0104 & \textbf{0.0808} & 0.3564 & \textbf{0.0276} & \textbf{4.4775} \\
\bottomrule
\end{tabular}
}
\end{table*}

\subsubsection{Qualitative Evaluation.}
As visually compared in Fig.~\ref{fig:qualitative_results} and Fig.~\ref{fig:prompt_vis}, existing approaches exhibit distinct limitations inherent to their generative mechanisms. Traditional GAN-based methods (e.g., CycleGAN, StegoGAN) often suffer from training instability and mode collapse, leading to over-smoothed textures and severe spectral distortions due to the limitations of adversarial objectives. As observed in the results, they fail to recover high-frequency details, producing blurry outputs that lack the photorealism required for optical imagery. On the other hand, LDM-based approaches like BBDM and cBBDM, while capable of generating sharper textures, struggle with the semantic ambiguity of SAR data. Without explicit semantic grounding, these models frequently misinterpret inherent speckle noise as structural features, resulting in geometric hallucinations and content misalignment. Similarly, the naive ControlNet, conditioned solely on raw SAR intensity, fails to filter out this noise, leading to ambiguous structural boundaries and visual artifacts rather than coherent objects. In contrast, OSCAR effectively bridges this modality gap. By leveraging the Optical-aware SAR Encoder to inject explicit semantic guidance—via Class-aware Text Prompts and Hierarchical Visual Prompts—our model successfully disambiguates valid structure from speckle noise. Consequently, OSCAR suppresses artifacts and preserves precise structural coherence, delivering photorealistic synthesis that faithfully aligns with the Ground Truth even in complex scenarios where baselines fail.

\subsubsection{Ablation Study.}
We evaluate the individual contributions of our core components on BigEarthNet-v2, as summarized in Table~\ref{tab:ablation_total}. In our experiments, the absence of cross-modal alignment (Aln.: --) indicates that the encoder was fine-tuned exclusively using SAR imagery without any optical-aligned distillation. Notably, we omit the results of the naive DINOv3-SAT as a control source. While it is a robust foundation model, its feature space is strictly optimized for optical reflectance. When SAR imagery is fed into the naive encoder, it produces "semantically shifted" features—misinterpreting SAR backscatter as unrelated optical patterns. These unaligned features act as misleading guidance (noise) for the ControlNet, triggering structural hallucinations and failing to provide the precise spatial-semantic constraints required for the diffusion process. Consequently, the naive model served as a lower bound in our tests, significantly underperforming even the unaligned configuration (i) which undergoes basic domain adaptation.

 Comparing configuration (i) and the Full Model (v) highlights the impact of cross-modal alignment. Although (i) achieves competitive FID and KID scores by focusing on matching raw textures, the Full Model (v) provides superior perceptual quality, leading in DISTS ($0.2404$), LPIPS ($0.3831$), and SAM ($0.1638$). This demonstrates that while an unaligned model can mimic simple patterns, our distillation strategy is essential to ensure the generated images follow the spectral and semantic logic of the optical domain.

\begin{figure*}[t]
    \centering
    \includegraphics[width=1.0\textwidth]{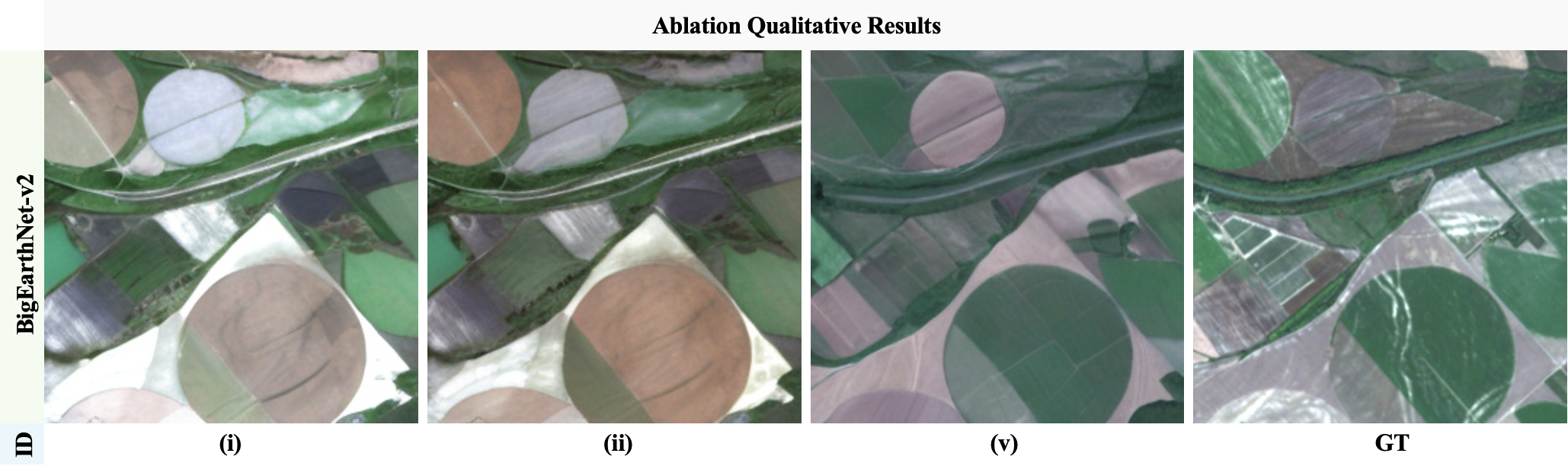}
    \caption{Qualitative comparison of ablation variants on BigEarthNet-v2.}
    \label{fig:ablation_qual}
\end{figure*}

We also observe a clear synergy between hierarchical and text guidance. Interestingly, using text prompts alone (iii) leads to a small drop in FID compared to (ii), likely because the model lacks the fine-grained spatial details needed to anchor the text-based textures correctly. However, when text and hierarchical prompts are combined in the Full Model (v), they work in harmony to reach the best overall balance and the highest quantitative results. This confirms that integrating detailed spatial guidance with global text context is key to resolving the inherent ambiguities in SAR data. As shown in the qualitative results in Fig.~\ref{fig:ablation_qual}, our Full Model produces images that are structurally accurate and exhibit color distributions much more consistent with real optical imagery, successfully reconstructing sharp boundaries while avoiding the structural drift seen in other configurations.

\subsection{Limitations and Discussion}
Despite the superior performance and semantic fidelity achieved by OSCAR, certain computational trade-offs exist. Primarily, the integration of a large-scale vision foundation model (DINOv3-SAT) and the latent diffusion backbone leads to a relatively high model weight and increased memory footprint compared to lightweight GAN-based architectures. Consequently, the iterative denoising process inherent in diffusion models results in slower inference speeds, which may limit real-time operational utility. However, these efficiency concerns are not prohibitive. The modular nature of our framework allows for the adoption of advanced sampling schedulers (e.g., DPM-Solver \cite{luDPMSolverFastODE2022}) or consistency-based distillation methods (e.g., LCMs \cite{luoLatentConsistencyModels2023}), which can effectively bridge the gap between generative quality and inference throughput.

Beyond computational aspects, there remain unexplored avenues for leveraging the rich metadata inherent in remote sensing datasets. Currently, OSCAR operates primarily on visual modalities; however, benchmarks like BigEarthNet-v2 and SEN12MS provide comprehensive spatiotemporal metadata, including latitude, longitude, and acquisition timestamps. Integrating these attributes as additional condition embeddings could significantly enhance the model's robustness against diverse seasonal and geographic variations. Furthermore, while our current framework utilizes image-level classification for semantic grounding, these datasets support pixel-level land cover annotations. Future iterations could replace the auxiliary classifier with a dense segmentation head to extract hierarchical prompts and class prompts. We anticipate that such granular semantic guidance would enable the reconstruction of even finer details and sharper boundary conditions, pushing the limits of S2O translation fidelity.

\section{Conclusion}
In this paper, we introduced OSCAR, a framework that redefines the boundaries of S2O translation by addressing the fundamental ill-posedness of the modality gap. While previous approaches often suffered from speckle-induced hallucinations and semantic drift, our method successfully mitigates these challenges by synergizing modality-aligned semantic priors with aleatoric uncertainty modeling. Through the integration of an optical-aware encoder and semantically-grounded generative guidance, OSCAR ensures that synthesized images are not only perceptually photorealistic but also topologically and semantically consistent with the underlying terrain. Extensive quantitative and qualitative evaluations across diverse global benchmarks (BigEarthNet-v2 and SEN12MS) demonstrate that our framework sets a new state-of-the-art, offering a robust and scalable solution for all-weather remote sensing interpretation. By effectively bridging the gap between noisy radar observations and high-fidelity optical scenes, OSCAR provides a reliable foundation for critical downstream earth observation tasks.

%
%
\bibliographystyle{splncs04}
\bibliography{references}

\setcounter{section}{0}
\setcounter{figure}{0}
\setcounter{table}{0}
\setcounter{equation}{0}
\renewcommand{\thesection}{S\arabic{section}}
\renewcommand{\thefigure}{S\arabic{figure}}
\renewcommand{\thetable}{S\arabic{table}}
\renewcommand{\theequation}{S\arabic{equation}}

\title{Supplementary Material for:\\ OSCAR: Optical-aware Semantic Control for Aleatoric Refinement in S2O Translation}

\author{
Hyunseo Lee\inst{1} \and
Sang Min Kim\inst{1} \and
Ho Kyung Shin\inst{1} \and
Taeheon Kim\inst{2} \and 
Woo-Jeoung Nam\inst{1}\thanks{Corresponding author.}
}

\authorrunning{H.-s. Lee et al.}

\institute{
\textsuperscript{1}Kyungpook National University \\
\textsuperscript{2}Korea Aerospace Research Institute \\
\email{
\{heart2002101, klasskoch, tlsghrud\}@knu.ac.kr, honey25@kari.re.kr, nwj0612@knu.ac.kr \\
}
}

\maketitle

\section{Detailed Experimental Setup}
\subsection{Data Preprocessing}
To ensure consistency across datasets, we implemented a unified pipeline targeting a spatial resolution of $256 \times 256$. For SEN12MS, native patches were used directly, whereas for BigEarthNet-v2, four adjacent $120 \times 120$ patches were merged and resized to preserve broader semantic context. SAR inputs were preprocessed using the Refined Lee Filter for speckle reduction and expanded to three channels $(VV, VH, VV-VH)$, while Sentinel-2 RGB bands served as the optical ground truth. Radiometrically, we applied channel-wise 3-sigma clipping to mitigate outliers, followed by min-max scaling to $[0, 1]$. During training, data augmentation included random horizontal and vertical flips, with final inputs normalized using dataset-specific mean and standard deviation.

Regarding the dataset partition, we randomly split both BigEarthNet-v2 and SEN12MS into training and testing sets with an 80:20 ratio. For the final evaluation, we constructed representative test subsets by randomly sampling 1,000 images from the BigEarthNet-v2 test set and 3,000 images from the SEN12MS test set.

\subsection{Implementation Details}
Our framework was implemented using PyTorch and trained on two NVIDIA GeForce RTX 3090 GPUs. We set the batch size to 8 and utilized gradient accumulation with 4 steps to ensure training stability. The model was optimized for 100k iterations using the Adam optimizer with a learning rate of $5 \times 10^{-5}$, employing a cosine annealing scheduler with a linear warmup of 100 steps. We utilized the pre-trained Stable Diffusion v2.1 as the backbone for our Latent Diffusion Model (LDM). For quantitative evaluation, all metrics were computed using the Denoising Diffusion Implicit Models (DDIM) sampler with 50 inference steps.

Regarding specific hyperparameters, the LoRA adapters for cross-modal alignment were configured with rank $r=8$, scaling factor $\alpha=16$, and dropout $p=0.05$. To stabilize the objective functions, we set the clipping thresholds for the class-balanced task loss to $\alpha_{min}=1.0$ and $\alpha_{max}=100.0$, the distillation temperature to $\mathcal{T}=4.0$, and the VicReg regularization weights to $\lambda_{inv}=25.0, \mu_{var}=25.0, \nu_{cov}=1.0$. During inference, the confidence threshold for class-aware prompting was set to $\tau=0.7$, $k=2$ and we applied Classifier-Free Guidance (CFG) with a scale of $s=5.5$, leveraging a null-token drop probability of 0.5 used during training.

\section{Additional Quantitative Comparisons}

\subsection{Evaluation of Semantic Alignment} To validate the proposed Cross-Modal Semantic Alignment, we first assess the discriminative capability of the Optical-aware SAR Encoder against the Optical Teacher. Since the Student is explicitly trained to internalize the Teacher's optical manifold, comparable classification performance serves as a direct proxy for successful domain alignment.

Table~\ref{tab:cls_results} shows the classification metrics on BigEarthNet-v2 and SEN12MS. As observed, the SAR Encoder achieves performance highly competitive with the Optical Teacher across all metrics (e.g., $AP^\mu$ gap within 2.8\%p). This confirms that our distillation strategy effectively transfers robust semantic priors from the optical domain to the SAR encoder, enabling it to extract modality-agnostic features despite relying solely on noisy SAR inputs.

\begin{table}[ht]
\caption{Classification performance comparison. The model trained via cross-modal distillation is marked with $\dagger$.}
\label{tab:cls_results}
\centering
\footnotesize
\setlength{\tabcolsep}{4pt}
\begin{tabularx}{\textwidth}{ll *{4}{>{\centering\arraybackslash}X}}
\toprule
\textbf{Dataset} & \textbf{Model} & $\mathbf{AP^M}$~$\uparrow$ & $\mathbf{AP^\mu}$~$\uparrow$ & $\mathbf{F_1^M}$~$\uparrow$ & $\mathbf{F_1^{\mu}}$~$\uparrow$ \\
\midrule
\multirow{2}{*}{BigEarthNet-v2} & Optical Teacher & 74.68 & 88.81 & 80.87 & 69.59 \\
& SAR Student$^\dagger$ & 71.35 & 86.00 & 77.17 & 64.98 \\
\midrule
\multirow{2}{*}{SEN12MS} & Optical Teacher & 80.61 & 93.90 & 75.34 & 87.41 \\
& SAR Student$^\dagger$ & 79.10 & 94.48 & 74.60 & 86.20 \\
\bottomrule
\end{tabularx}
\end{table}

\section{Additional Qualitative Comparisons}

\subsection{Qualitative Analysis of Cross-Modal Feature Alignment}
To validate the effectiveness of our distillation strategy, we visualize the internal representations of the SAR Student compared to the Optical Teacher. We focus on two complementary perspectives: (1) global semantic consistency via Principal Component Analysis (PCA) of patch tokens, and (2) local discriminative focus via Attention Maps.

\begin{figure}[t]
    \centering
    \includegraphics[width=1.0\textwidth]{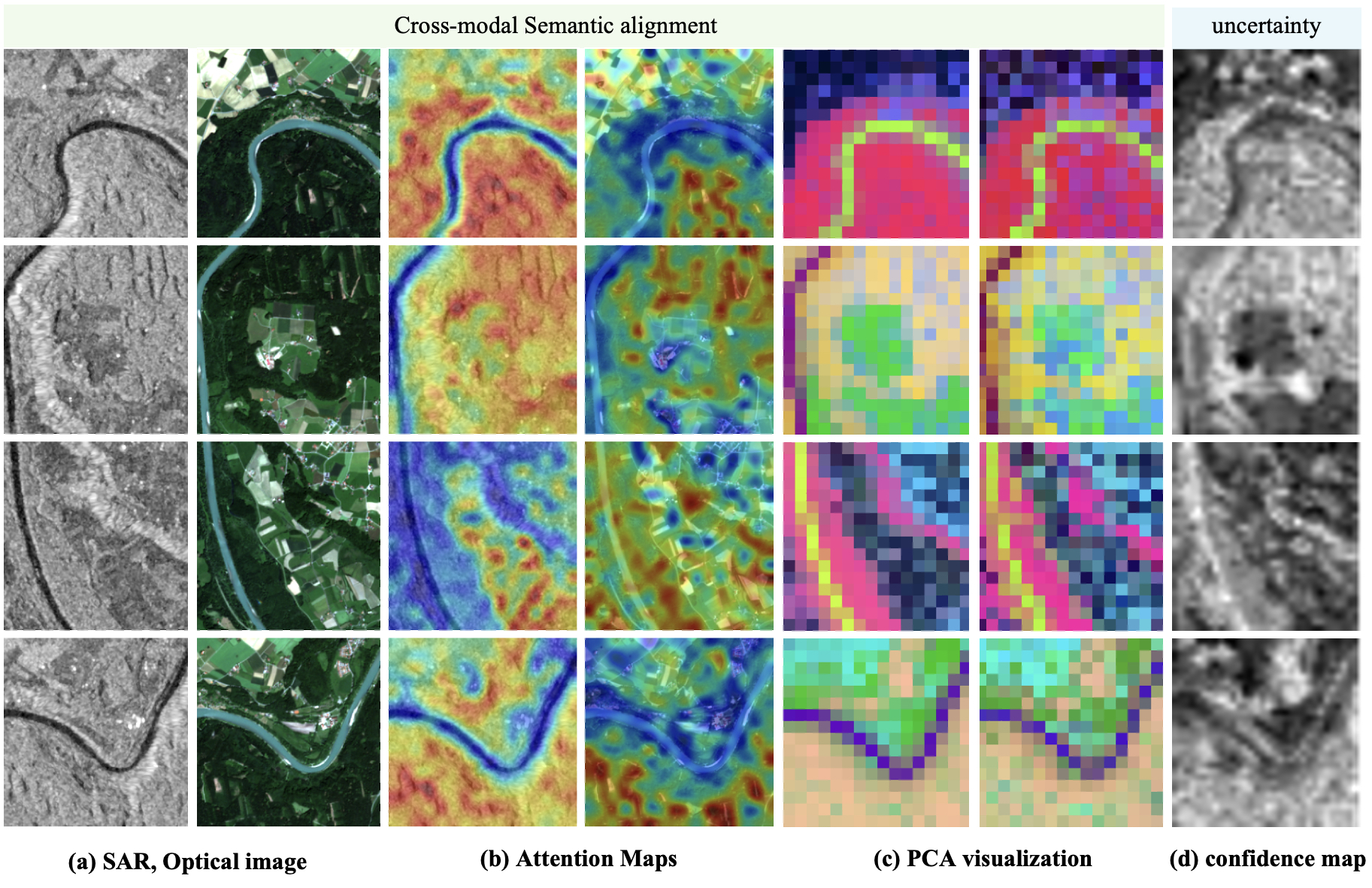}
    \caption{Visualization of Cross-Modal Feature Alignment and Aleatoric Confidence in BigEarthNet-v2
    (a) Columns 1-2: Input SAR and reference Optical images.
    (b) Columns 3-4: Attention maps of SAR Student and Optical Teacher.
    (c) Columns 5-6: PCA visualization of feature representations for the Student and Teacher.
    (d) Column 7: The estimated confidence map utilized in the objective of our Semantically-grounded ControlNet.}
    \label{fig:feature_vis}
\end{figure}

As shown in Fig.~\ref{fig:feature_vis} (a)$\sim$(c), we extract features from the last transformer block of the DINOv3-SAT backbone. The PCA visualization projects high-dimensional features into an RGB space, where color similarity implies semantic affinity. Remarkably, the SAR Student exhibits a semantic layout nearly identical to that of the Optical Teacher. It successfully delineates distinct land covers (e.g., separating water bodies from vegetation) and maintains sharp boundaries, demonstrating that the student has learned to suppress inherent SAR speckle noise to recover clean semantic structures.

Furthermore, we analyze the attention maps of the [CLS] token with respect to the patch tokens. While the Optical Teacher naturally attends to salient objects within the clean optical domain, the SAR Student mirrors this behavior, focusing on the same semantic regions. This strong spatial correspondence confirms that our Cross-Modal Semantic Alignment effectively transfers the discriminative capability of the Optical domain to the SAR encoder, enabling it to serve as a robust semantic guide.

\begin{figure}[t]
    \centering
    \includegraphics[width=1.0\textwidth]{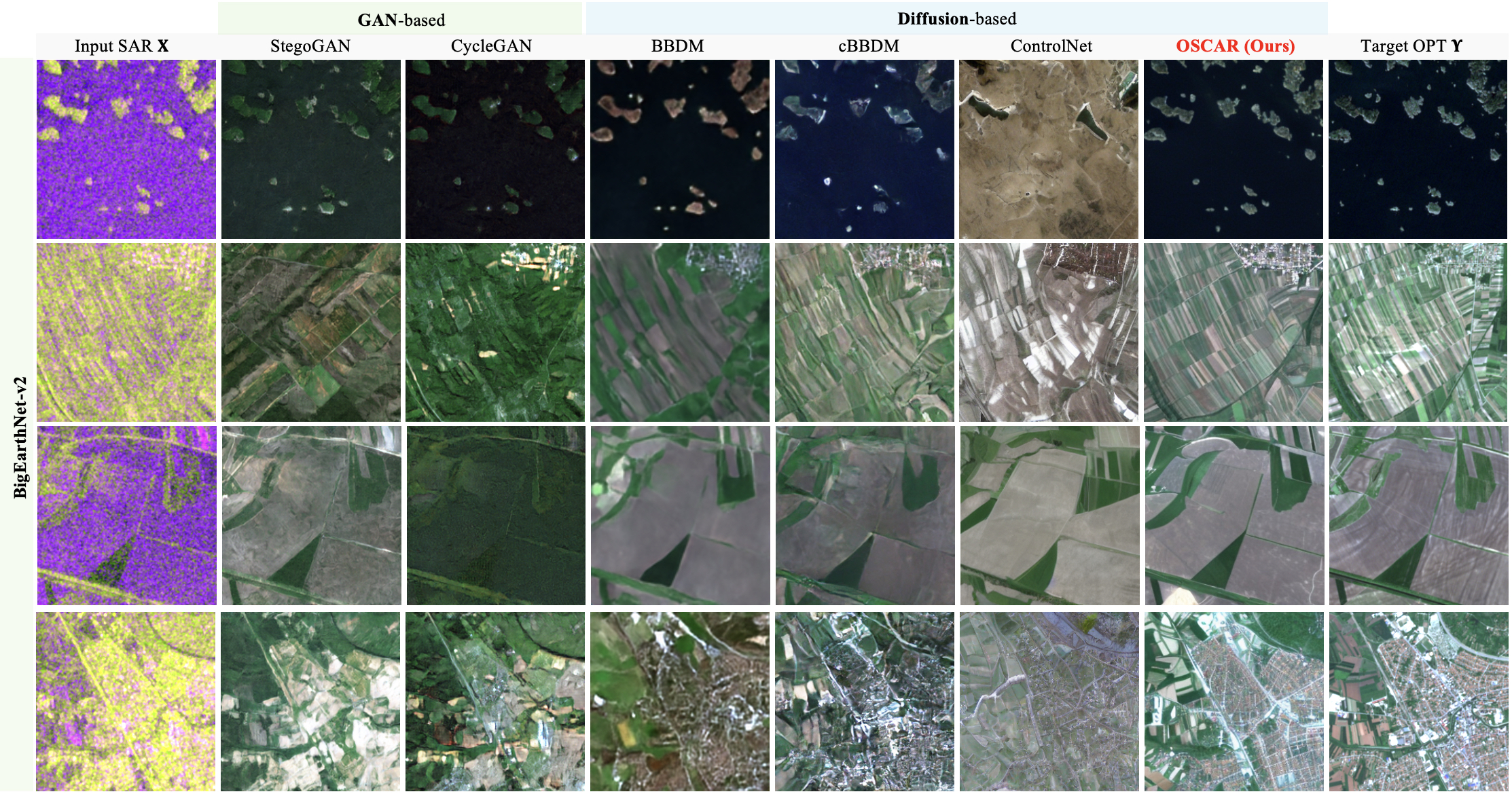} 
    \caption{Qualitative comparison on the BigEarthNet-v2 dataset. Each row represents a different sample. The columns display, from left to right: Input SAR X, results from GAN-based methods (StegoGAN, CycleGAN), results from Diffusion-based methods (BBDM, cBBDM, ControlNet, OSCAR (Ours)), and the Target OPT Y (Ground Truth).}
    \label{fig:more_results_bigearth}
\end{figure}

\subsection{Extended Visual Results on SAR-to-Optical Translation}
\begin{figure}[t]
    \centering
    \includegraphics[width=1.0\textwidth]{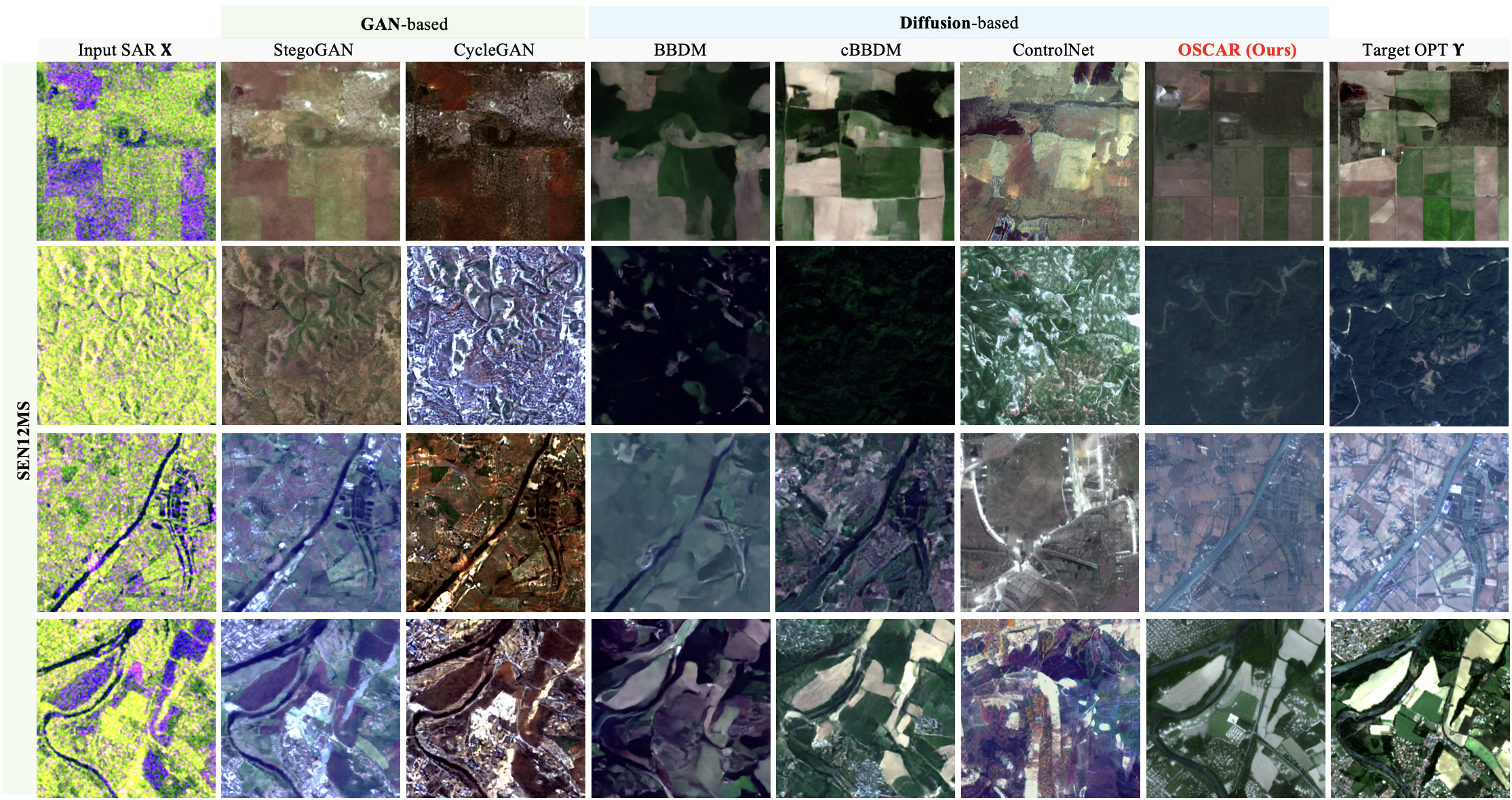}
    \caption{Qualitative comparison on the SEN12MS dataset. Each row represents a different sample. The columns display, from left to right: Input SAR X, results from GAN-based methods (StegoGAN, CycleGAN), results from Diffusion-based methods (BBDM, cBBDM, ControlNet, OSCAR (Ours)), and the Target OPT Y (Ground Truth).}
    \label{fig:more_results_sen12ms}
\end{figure}

We present additional qualitative results to further demonstrate the generalization capability and visual fidelity of OSCAR. Fig.~\ref{fig:more_results_bigearth} and Fig.~\ref{fig:more_results_sen12ms} showcase randomly selected translation samples from the BigEarthNet-v2 and SEN12MS test sets, respectively.

As illustrated, our method consistently generates photorealistic optical imagery across diverse landscapes, ranging from dense urban structures and agricultural fields to complex waterways. Crucially, OSCAR effectively suppresses the inherent speckle noise of SAR inputs while preserving fine-grained structural details. The generated images exhibit plausible color distributions and realistic textures that closely align with the ground truth, validating the robustness of our optical-aware semantic control even in challenging scenarios.
\subsubsection{Visualization of Aleatoric Confidence.}
In addition to feature alignment, we visualize the spatial confidence maps used to modulate the control signal, as shown in Fig.~\ref{fig:feature_vis} (d). We specifically capture this visualization at the intermediate timestep $t=T/2$ during the training of our Semantically-grounded ControlNet. This phase is critical as the model transitions from defining coarse global layouts to refining local textures.

The map quantifies the pixel-wise reliability of the extracted semantic features. Brighter regions indicate higher confidence (i.e., low aleatoric uncertainty), typically corresponding to distinct structural boundaries where the SAR signal offers unambiguous cues. Conversely, darker regions represent lower confidence, often associated with homogeneous areas degraded by speckle noise. By leveraging this map at crucial timesteps like $t=T/2$, the model learns to adaptively prioritize robust semantic features while attenuating the influence of uncertain representations.

\end{document}